\documentclass[letterpaper, 10 pt, conference]{ieeeconf}  % Comment this line out if you need a4paper

\IEEEoverridecommandlockouts                              % This command is only needed if 
\usepackage{graphicx} % for pdf, bitmapped graphics files
\usepackage{epsfig} % for postscript graphics files
\usepackage{amsmath}
\usepackage{amsfonts}
\usepackage{amssymb}
\usepackage{siunitx}
\usepackage{algorithm}
\usepackage{algpseudocode}
\usepackage{subcaption}
\usepackage{xcolor}
\usepackage{todonotes}

\title{\LARGE \bf
Aerial Inspection Behaviors via RL-based Quadrotor Control for Under-canopy Forest Environments  
}

\author{Fausto Mauricio Lagos Suarez$^{1}$, Akshit Saradagi, Vidya Sumathy, Viswa Narayanan Sankaranarayanan \\ and George Nikolakopoulos% <-this % stops a space
\thanks{*This work has been funded by the European Union's Horizon Europe Research and Innovation Program, under the Grant Agreement No. 101119774 SPEAR.}% <-this % stops a space
\thanks{*This research was conducted using the resources of High Performance Computing Center North (HPC2N). Additionally, the RL-training were enabled by resources provided by the National Academic Infrastructure for Supercomputing in Sweden (NAISS), partially funded by the Swedish Research Council through grant agreement no. 2022-06725.}
\thanks{$^{1}$Fausto Lagos is the corresponding author of the article
        {\tt\small faulag@ltu.se}. The authors are with the Robotics and AI group, in the Department of Computer Science, Electrical and Space Engineering at Luleå University of Technology, Sweden.}%
}

\begin{document}

\maketitle
\thispagestyle{empty}
\pagestyle{empty}

%%%%%%%%%%%%%%%%%%%%%%%%%%%%%%%%%%%%%%%%%%%%%%%%%%%%%%%%%%%%%%%%%%%%%%%%%%%%%%%%
\begin{abstract}
This paper addresses the problem of using a deep Reinforcement Learning (RL)-based low-level Quadrotor controller within an autonomous Quadrotor navigation stack for aerial inspection missions in under-canopy forest environments. Specifically, the article presents an end-to-end (mapping states to RPMs) Quadrotor control policy that achieves inspection view-pose tracking (simultaneous position and yaw reference tracking), which is crucial for various target inspection behaviors and point-to-point navigation in forests. To ensure safe and reliable deployment of the end-to-end RL controller in long-range missions, this article utilizes a higher navigation guidance layer comprising of a Traveling Salesman Problem planner (TSP) and a Rapidly-exploring Random Tree Star (RRT*) planner. Over a known map of a forest and a set of user-specified inspection regions, the TSP planner finds the optimal visitation sequence. Between two target regions, collision-free paths that respect the tracking limitations of the lower end-to-end RL policy are generated by an RRT* planner. Through five target inspection scenarios, this article demonstrates that an RL-based motor-level stabilizing controller, supported by a navigation guidance layer, can be used effectively as the low-level inspection execution module for under-canopy forest inspection missions.
\end{abstract}

%%%%%%%%%%%%%%%%%%%%%%%%%%%%%%%%%%%%%%%%%%%%%%%%%%%%%%%%%%%%%%%%%%%%%%%%%%%%%%%%
\section{INTRODUCTION}

% Undercanopy Navigation and Inspection in Forests. How is it different from navigation in other environments?

% Why aerial robots for forest inspection? 
Quadrotors are attractive platforms for autonomous inspection and navigation because they can hover, maneuver in confined spaces, and operate in cluttered environments where ground robots or larger aerial systems may be less suitable \cite{aerial_workers}. To enable autonomous aerial inspection, however, the vehicle must combine inspection-oriented motion behavior with reliable navigation in cluttered environments. In conventional autonomy pipelines, safe flight is usually achieved by combining a motion planner with a structured control stack \cite{yaw_forest, shi_multi_waypoint}. On the control side, representative classical works include Proportional Integral and Derivative control (PID) \cite{Lopez_PID}, and geometric tracking control on $SE(3)$ \cite{gamagedara_geometric}. On the planning side, sampling-based methods such as RRT* remain a standard choice for generating collision-free paths with asymptotic optimality guarantees \cite{gao_rrt}. These classical approaches are highly effective and remain strong baselines for quadrotor stabilization and tracking. Accordingly, the motivation for this work is whether a learned motor-level controller can serve as a practical alternative execution layer inside a planner-guided autonomy stack. This question is relevant because conventional pipelines often rely on an explicit decomposition into planning, outer-loop position control, inner-loop attitude control, and control allocation, whereas an end-to-end policy can, in principle, map tracking-relevant state errors directly to motor commands. Such a formulation may reduce controller hand-design while preserving the higher-level planner that provides collision-free references \cite{song_optimal_vs_rl}. 

Research on low-level RL for quadrotors has progressed substantially over the last several years. Early work showed that a neural policy could map state directly to actuator commands for quadrotor stabilization \cite{hwangbo_control_2017}, and subsequent studies extended this direction to faster training and effective sim-to-real transfer \cite{Eschmann_learning_to_fly}. Additional results have pushed RL-based aerial control toward high-speed autonomous flight, and champion-level drone racing \cite{kaufmann_champion-level_2023}. At the same time, recent literature shows that learning-based autonomy is diversifying rather than converging to a single paradigm. Published work has explored multi-agent RL for low-level control of quadrotors, separating the position, roll, and pitch tracking from the yaw tracking as in classical cascade approaches \cite{yu_multi_RL_low_level_control}, RL policies capable of handling multiple aerial tasks like stabilization, autonomous race, and random velocities tracking \cite{xing_multi_task}, and RL for vision-based autonomous navigation in cluttering environments with speed adaptation \cite{zhao_speed_adaptation, yu_mavrl}, and for inspection path planning in industrial like scenarios \cite{Malczyk_semantically}. These results are important because they suggest that learned aerial control is no longer limited to hover demonstrations. Compared with other systems questions, less attention has been paid to the highly relevant problem addressed by this work.

The main contributions of this work are threefold. First, we present an end-to-end RL-based low-level quadrotor controller that maps state observations directly to motor RPM commands for inspection view-pose tracking, namely simultaneous position and yaw reference tracking. Second, we demonstrate that this learned controller can be reliably deployed as the low-level execution module within a higher-level navigation-guidance framework for long-range under-canopy forest-inspection missions. Third, we integrate the controller with a TSP-based target-visitation planner and an RRT*-based collision-free path planner over a known forest map and user-defined inspection regions, and we validate the complete system across five inspection scenarios. 

% The rest of the paper is organized as follows: Section \ref{sec:problem_setup} elaborates on the assumptions and problem statement of interest for this work. Section \ref{sec:methodology} presents details on the autonomy architecture designed for this work, the design of the proposed low-level RL-based quadrotor controller, and the training process. The inspection scenarios and results of the validation are presented in Section \ref{sec:validation}. Finally, Section \ref{sec:conclusions} concludes the paper and proposes future steps to continue research on the topic of interest to this work.
%
\section{Problem setup} \label{sec:problem_setup}
%
%\subsection{Problem formulation}

% Define the problem. Assumptions, Expectations 

In designing target inspection missions in under-canopy forest environments, this article assumes that a map of the environment is known a priori (built from exploration missions or derived from a digital twin). The focus of the article is not on reactive navigation using onboard perception from cameras and Lidar, but on designing a two-level navigation pipeline with a higher safe path planning layer (over a known map) and a lower RL-based path tracking and inspection behavior layer, assuming odometry derived from the fusion of IMU and GNSS.  

This article considers typical low-level aerial behaviors necessary in long range inspection missions in forests, such as point-to-point navigation, inspection pose tracking, view scanning, and tracking circular and spiral inspection trajectories. In this article, the inspection behaviors are derived from a base RL policy capable of tracking inspection view poses, wherein position and yaw references are simultaneously tracked, in the presence of state measurement noise and external disturbances. The RL policy aims to track a time-indexed sequence of reference poses $r_t = (p_t^\ast, \psi_t^\ast)$, where $p_t^\ast \in \mathbb{R}^3$ is the desired position at time $t$ and $\psi_t^\ast \in [-180^\circ, 180^\circ)^1$ is the desired yaw attitude.

Assuming a known map and odometry derived from the fusion of IMU and GNSS and no other external sensing, this article explores the design of a higher navigation guidance layer that effectively solves the macro-level optimal route finding problem and local obstacle avoidance. Such a guidance layer is expected to supply a sequence of inspection view poses to the lower RL policy. The quadrotor, therefore, operates within a hierarchical autonomy architecture in which a TSP high-level planner creates the optimal sequence of targets and an RRT*-based planner provides collision-free paths. 

%The quadrotor with state $\mathcal{S}$ evolves under nonlinear rigid-body dynamics and is controlled through the four rotor angular speeds. 
%

\section{Methodology} \label{sec:methodology}
\subsection{Autonomy Architecture}
\begin{figure}[ht]
    \centering
    \includegraphics[width=0.48\textwidth]{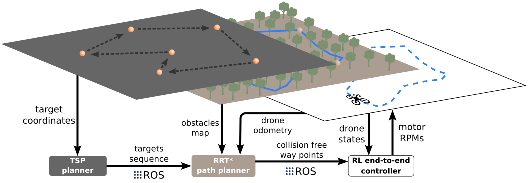}
    \caption{TSP-RRT*-RL navigation stack architecture.}
    \label{fig:stack_architecture}
\end{figure}

Figure \ref{fig:stack_architecture} illustrates the hierarchical navigation architecture used in this work. At the highest level, the TSP planner receives the set of inspection target coordinates and computes an ordered target sequence. This sequence is then passed to the RRT* path planner, which also receives the obstacle map and the current drone odometry to generate collision-free waypoints between successive targets. Finally, these waypoint references, together with the current drone states, are provided to the RL end-to-end controller, which outputs the four motor RPM commands required to track the planned path. In this stack, target ordering is handled by the TSP layer, collision avoidance is handled by the RRT* planner, and low-level path execution is handled by the learned motor-level controller.
\subsection{Navigation Guidance}
\subsubsection{User interface and TSP-based large-scale planning}
The TSP planner is designed to take the target inspection poses as input and generate every single possible tour in which the inspection targets could be visited, starting at the current location of the quadrotor and finishing the tour at the same location. The path cost is calculated using the Euclidean distance between two targets without considering altitude $z$ or yaw ($\psi$) rotation. The optimal tour outcome of the TSP-planner corresponds to the tour for which the total distance is an absolute minimum. Then, the first target of the optimal tour is published to the RRT* planner, and a new target is published once the current quadrotor odometry is within one meter of the previous target.
\subsubsection{Point-to-point Local Navigation with RRT* Planner}
The planner is implemented as a 3D informed RRT* module embedded in a ROS2 node, where the quadrotor odometry is used as the start state and each incoming target position defines a new planning query. The algorithm incrementally expands a tree in Cartesian space by sampling collision-free candidate notes within bounded workspace limits, steering with a fixed step size, selecting the minimum-cost parent among nearby nodes, and rewiring local branches to improve path length. Collision avoidance is enforced during done expansion, parent selection, rewiring, and path smoothing through point -and segment-based checks against vertically bounded cylindrical obstacles enlarged by an inflation radius, thereby guaranteeing that only feasible path segments are inserted into the tree. Once a valid solution is found, the planner reconstructs and smooths the waypoint sequence, computes a heading profile from the local path direction, publishes the global path for visualization, and generates successive reference setpoints for downstream tracking control. The RRT* planner is aware of the constraints established by the training initial conditions and failure termination conditions when it creates the sequence of setpoints to track.
\subsection{RL-based inspection inner loop controller}

\subsubsection{Expectations from the RL policy}
Given a planner-generated sequence of collision-free waypoint references $\{\left.(p_t^\ast, \psi_t^\ast)\right|_{t=0}^T\}$, the problem addressed in this work is to determine whether a motor-level RL policy trained primarily for robust hover recovery can be reused as the low-level execution module of a planner-guided navigation pipeline for forest inspection. Formally, the goal is to synthesize a feedback law $a_t = \pi_\theta(\mathbf{o}_t)$ such that the closed-loop system remains stable and drives the tracking error toward zero along the executed path, i.e., $e_{p, t} = ||p_t - p_t^\ast||_2^2 \rightarrow 0$, and $e_{\theta, t} = \theta_t - \theta_t^\ast \rightarrow 0$, where $\theta$ is the geodesic angle as explained in section \ref{sec:reward}, while keeping the remaining translational and rotational states bounded throughout the maneuver.  

\subsubsection{RL setup for aerial inspection view pose tracking} This work proposes an RL-based inner loop controller modeled as a Partial Observable Markov Decision Process (POMDP) represented as a tuple of five elements $(\mathcal{S}, \mathcal{A}, \mathcal{P}, \mathcal{R}, \gamma)$, where $\mathcal{S}$ is the state space, $\mathcal{A}$ is the action space, $\mathcal{P}$ is the state transition probability, $\mathcal{R}$ is the reward function, and $\gamma \in (0, 1)$ is the discounting factor. 
\paragraph{Observation space}
The observation space of the agent is defined as a vector $\vec{\mathbf{o}}_t = [\begin{array}{ccccc}\tilde{\mathbf{e}}_{p, t} & \tilde{\mathbf{q}}_{e, t} & \tilde{\mathbf{v}_t} & \tilde{\omega}_t & \mathbf{a}_{t-1}\end{array}]^T \in \mathbb{R}^{17}$, where $\tilde{\mathbf{e}}_{p, t} \in \mathbb{R}^3$ is the position error with noise, $\tilde{\mathbf{q}}_{e, t} \in \mathbb{R}^4$ is the error quaternion with noise, $\tilde{\mathbf{v}_t} \in \mathbb{R}^3$ is the linear velocity with noise, is the $\tilde{\omega}_t \in \mathbb{R}^3$ the angular velocity with noise, and $\mathbf{a}_{t-1} \in [-1, 1]^4$ is the last policy action required to keep the observation Markov due the motor RPM used as action space. We add Gaussian noise to all components of the observation space, except the last four actions, to improve the policy's robustness to real-time sensor noise, which is essential for real-time implementation. The magnitude of the added noise is as specified in Table \ref{tab:observation_noise}.
\renewcommand{\tabcolsep}{1mm}
\renewcommand{\arraystretch}{1.5}
\renewcommand{\arrayrulewidth}{1pt}
\begin{table}[h!]
    \centering
    \begin{tabular}{cc}
        \hline
        \textbf{Reward Component} & \textbf{Noise magnitude} \\
        \hline
        $\tilde{\mathbf{e}}_{p, t}$ & $\mathcal{N}(0,\sigma_p^2I_3)$, $\sigma_p = 10^{-3}$ \\
        $\tilde{\mathbf{q}}_{e, t}$ & $\mathcal{N}(0, \sigma_q^2I_4), \sigma_q = \num{2e-3}$ \\
        $\tilde{\mathbf{v}_t}$ & $\mathcal{N}(0, \sigma_v^2I_3), \sigma_v = 10^{-3}$ \\
        $\tilde{\omega}_t$ & $\mathcal{N}(0, \sigma_\omega^2I_3), \sigma_\omega = \num{2e-3}$ \\
        \hline
    \end{tabular}
    \caption{Magnitude of the Gaussian noise injected in each state component of the observation space.}
    \label{tab:observation_noise}
\end{table}
\paragraph{Acion space}
This article aims to develop an end-to-end policy that directly maps the drone states to the raw control inputs. The action space is a vector $\vec{a} \in [-1, 1]^4$ of the normalized motor RPM values mapped in the simulation to $\text{RPM}\in (0, 65535)^4$ using the linear approximation $RPM_i = \left.RPM_{Hover} \cdot \left(1 + \frac{a_i}{2}\right)\right|_{i=1}^4$
where $RPM_{Hover} = \sqrt{\frac{gm}{4k_f}}$, with $g$ the gravitational acceleration magnitude, $m$ the drone mass, and $k_f$ the torque constant. We use motor RPMs as the action space because of their convenience in smooth and precise control of the quadrotor toward the desired pose \cite{Eschmann_learning_to_fly, zhang_rpms}.
\paragraph{Reward shaping} \label{sec:reward}
In an RL architecture, the reward term is one of the most important elements as it indirectly guides the policy to learn the desired logic. The goal of RL algorithms is to train an agent to obtain the most cumulative reward over time. 
In this paper, a multi-component reward function is designed to achieve safe performance in navigation and forest inspection scenarios. In this article, we propose the following reward function:
\begin{multline}
    R(t) = s^R + 0.25\mathbf{e}_{xy}^R + 0.25\mathbf{e}_z^R + 0.15\mathbf{v}^R + 0.2\theta^R - 0.02\mathbf{a}_\Delta^P,
    \label{eq:reward_function}
\end{multline}
where the individual components are as defined below.   
\textbf{Survival reward:} We use $s^R = 0.01$ as the survival reward to incentivize the agent to maintain its life, i.e., don't crash or terminate the learning episode.
\textbf{Position error reward ($\mathbf{e}_{xy}^R + \mathbf{e}_z^R$):} It is defined as a dual-bandwidth exponential reward used to incentivize the agent to achieve the target with high precision. The position error reward is split into two components: $\mathbf{e}_{xy}$ and $\mathbf{e}_z$ to allow the policy to learn how to decouple the altitude from the horizontal position.
The position error reward is defined as:
\begin{align} \label{eq:position_error_reward}
    \mathbf{e}_{xy}^R &= 0.6 e^{-4||(x,y)_t - (x,y)^\ast||^2} + 0.4 e^{-150||(x,y)_t - (x,y)^\ast||^2}, \\
    \mathbf{e}_{z}^R &= 0.6 e^{-4||z_t - z^\ast||^2} + 0.4 e^{-150||z_t - z^\ast||^2}.
\end{align}
\textbf{Linear velocity reward ($\mathbf{v}^R$):} Rewards the agent progressively while the linear velocity magnitude tends to zero. The linear velocity reward is defined as
\begin{equation} \label{eq:linear_velocity_reward}
    \mathbf{v}^R = e^{-1.5||v||^2},
\end{equation}
\textbf{Geodesic angle reward ($\theta^R$):} The geodesic angle represents the shortest angular distance between two 3D orientations (the target orientation and the drone's current orientation). We use the axis-angle representation of the error quaternion to calculate the geodesic angle as $\theta = 2\arctan2(||v||, w)$, where $v$ is the vector part, and $w$ the scalar part of the error quaternion. The geodesic angle reward is defined as:
\begin{equation} \label{eq:geodesic_angle_reward}
    \theta^R = 0.6\cdot e^{-0.5\theta^2} + 0.4\cdot e^{-150.0\theta^2},
\end{equation}
\textbf{Smoothness penalty ($\mathbf{a}_\Delta^P$):} Defined as the weighted squared Euclidean norm of the difference between consecutive actions:
\begin{equation} \label{eq:smoothness_penalty}
    \mathbf{a}_\Delta^P = ||a_t - a_{t - 1}||_2^2,
\end{equation}
is used to add damping and encourages the agent to approaching the target slowly, reducing the overshoot response.

\textbf{Remark:} In inspection-like scenarios, the controller must track any rotational yaw angle. To allow the policy to understand the environment state, the drone states ($\tilde{\mathbf{e}_{p,t}}$, $\tilde{\mathbf{v}_t}$, and $\tilde{\mathbf{\omega}_t}$) are rotated to the body coordinate frame ($\mathcal{B}$) before set the observation space. At the same time, the reward function is still calculated in the inertial coordinate frame ($\mathcal{I}$). This is beneficial to the policy to learn faster to control the drone's body, while the reward function acts as an external judge of the absolute physical realities, enabling the policy to learn how to track any target yaw with high accuracy, which is a desired behavior in inspection tasks. 
\paragraph{Neural network}
The actor (policy) neural network is a Multi-layered Perceptron (MLP) network with four layers: the first is a layer of 17 inputs (the observation space), the two hidden layers have 64 fully connected nodes with $\tanh$ activation functions, and the last layer is a four-node layer (the action space). The critic (value function) network has the same architecture, except for the last layer, which is a one-node layer yielding the output of the value function.
%
%\section{RL-training and design of inspection behaviors}
\subsection{Training}
For the training process, we used our own training framework built on top of Gym-PyBullet-Drones \cite{pybullet_drones}, a gymnasium environment that uses PyBullet Physics as the physics engine and Stable-Baselines3 as the library of reliable implementations of RL algorithms in PyTorch. Our training framework inherits the base training environments from Gym-PyBullet-Drones, adds new artifacts to enable ROS2 communication across navigation stack components, along with simulation enhancements for the specific forest scenarios used for testing. The training was executed on one CPU with four parallel environments running on a Compute-skylake Kebnekaise HPC2N node with an Intel Xeon Gold 6132 CPU and 192 GB RAM.
\begin{figure}[h!]
    \centering
    \includegraphics[width=0.8\linewidth]{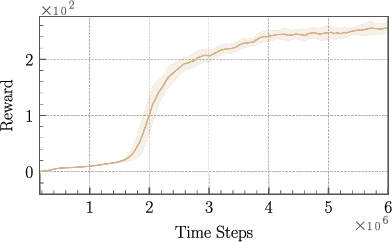}
    \caption{Evolution of the episode rollout reward.}
    \label{fig:learning}
\end{figure}

The physical parameters used to simulate the Crazyflie 2.1 in $\times$ configuration are explicitly shown in Table \ref{tab:crazyflie_parameters}. The control frequency corresponding to a complete interaction loop between the agent and the environment is 100 Hz.
\begin{table}[h!]
    \centering
    \begin{tabular}{ccc}
        \hline
        \textbf{Parameter} & \textbf{Notation} & \textbf{Value} \\
        \hline
        Mass &$m$ & 0.033 [$Kg$] \\
        Arm length &$d$ & \num{39.73d-3} [$m$] \\
        Torque constant & $k_f$ & \num{3.16d-10} \\
        Moment constant & $k_m$ & \num{7.49d-12} \\
        Propeller radius &$p$ & \num{23.1348d-3} [$m$] \\
         \hline
    \end{tabular}
    \caption{Physical parameters of the Crazyflie 2.1 in $\times$ configuration.}
    \label{tab:crazyflie_parameters}
\end{table}
\section{Validation setup and Results} \label{sec:validation}
\subsection{Waypoint tracking from a stabilization policy}
As the policy is trained with random starting positions within a cylinder of radius 2 m and height 2 m around the target position at $(x, y, z)^\ast = (0, 0, 1)$, and a tolerance exploration position error of 3 m around the target is allowed through the failure termination conditions, to ensure safety navigation, the sequence of waypoints must be created considering the position of the next waypoint into a cylinder of radius $r < 3$ m. The RRT* is aware of that boundary to calculate the waypoints path.
\begin{figure}[h!]
    \centering
    \includegraphics[width=0.65\linewidth]{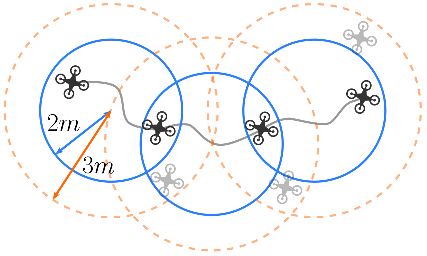}
    \caption{Constraints on the definition of the waypoint sequence. In blue, the constraints related to the initialization during training, in yellow, the ones defined as termination conditions.}
    \label{fig:waypoint}
\end{figure}
\subsection{Inspection scenarios for testing the stabilization policy}
\subsubsection{Under canopy forest navigation}
\begin{figure*}[h!]
    \centering
    \begin{subfigure}[t]{0.33\linewidth}
        \centering
        \includegraphics[width=0.9\textwidth]{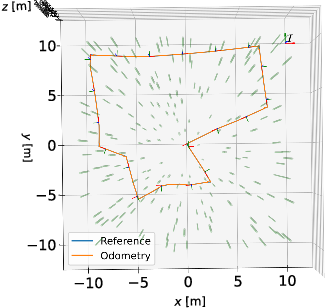}
    \end{subfigure}
    \hfill
    \begin{subfigure}[t]{0.62\linewidth}
        \centering
        \includegraphics[width=0.9\textwidth]{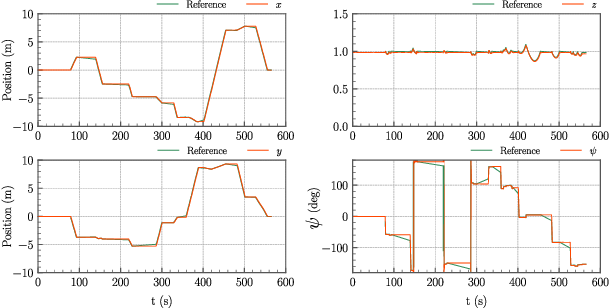}
    \end{subfigure}
    \caption{Top view (left) and detailed trajectories with references in the under canopy navigation test. The user defines 8 reachable inspection locations, the TSP planner computes the optimal sequence of targets, and the RRT* creates the collision-free path to be tracked.}
    \label{fig:inspection_mission}
\end{figure*}

We designed a forest-like scenario, where the under-canopy trunks are abstracted as cylinders of random radius $r \in (0.15, 0.25)$ m and constant height $h = 2$ m. The scenario corresponds to a rectangular area of 20 $m^2$ with 200 under-canopy trunks randomly located and with a random separation between trunks $d \in (1, 2)$ m. Eight inspection target locations are given to the TSP planner to calculate the optimal target sequence and published to the RRT* planner to create the collision-free waypoint path that the RL-based low-level controller must track. Figure \ref{fig:inspection_mission} presents the performance of the RL-based low-level controller in tracking the collision-free waypoints calculated by the RRT* planner.
\subsubsection{Tracking a sequence of view poses}
\begin{figure*}[h!]
    \centering
    \begin{subfigure}[t]{0.33\linewidth}
        \centering
        \includegraphics[width=0.8\textwidth]{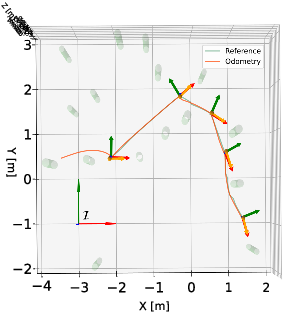}
    \end{subfigure}
    \hfill
    \begin{subfigure}[t]{0.62\linewidth}
        \centering
        \includegraphics[width=0.9\textwidth]{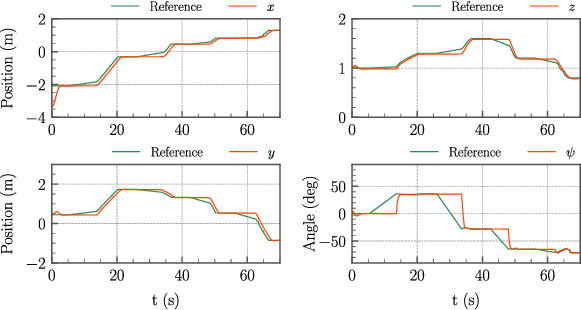}
    \end{subfigure}
    \caption{View of the desired inspection poses (left) and the references calculated by the RRT* planner for collision-free navigation among trees and the actual trajectories followed by the quadrotor.}
    \label{fig:inspection_poses}
\end{figure*}

In one inspection area of interest, the quadrotor is required to follow a collision-free path and inspect five trees at varying altitudes and a safety distance from the trunk of 1 m. The quadrotor must reach each target tree, tracking the desired view pose, pointing its positive $x$ axis toward the three. Figure \ref{fig:inspection_poses} shows the completed inspection task; the drone body frame overlaps the inspection pose (orange arrow) in each tree inspection location.
\subsubsection{Scene scanning at a desired position}
\begin{figure*}[h!]
    \centering
    \begin{subfigure}[t]{0.33\linewidth}
        \centering
        \includegraphics[width=0.8\textwidth]{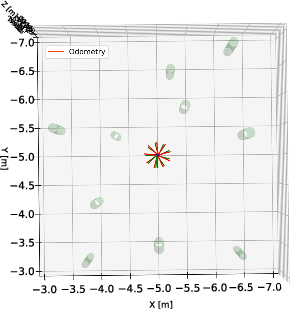}
    \end{subfigure}
    \hfill
    \begin{subfigure}[t]{0.62\linewidth}
        \centering
        \includegraphics[width=0.9\textwidth]{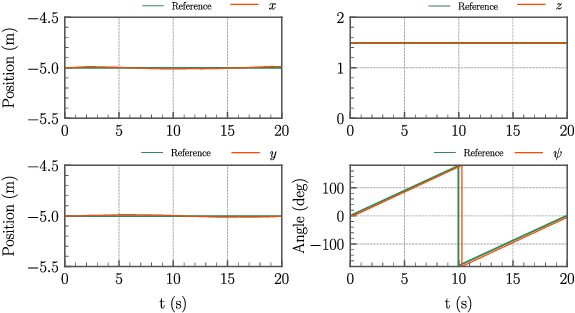}
    \end{subfigure}
    \caption{Scanning of a specific area/scene in stationary flight while rotation $360^\circ$. The quadrotor maintains its position with high accuracy w.r.t. the reference.}
    \label{fig:inspection_stationary_rotation}
\end{figure*}

Once the quadrotor reaches the desired inspection area, it must slowly rotate $360^\circ$ maintaining its altitude. Figure \ref{fig:inspection_stationary_rotation} illustrates the performance of the controller in this specific scenario.
\subsubsection{Track the circumference of a tree at a specific height}
\begin{figure*}[h!]
    \centering
    \begin{subfigure}[t]{0.33\linewidth}
        \centering
        \includegraphics[width=0.8\textwidth]{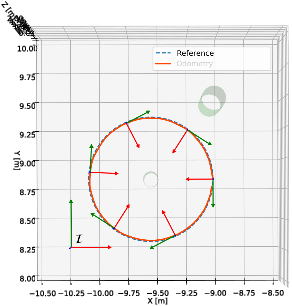}
    \end{subfigure}
    \hfill
    \begin{subfigure}[t]{0.62\linewidth}
        \centering
        \includegraphics[width=0.9\textwidth]{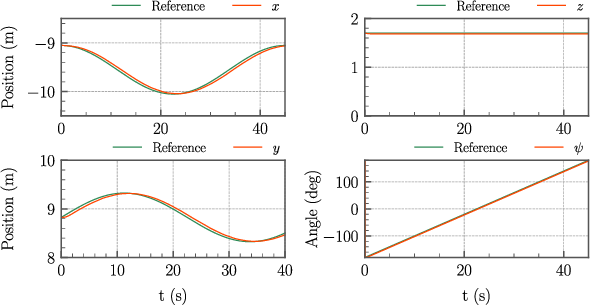}
    \end{subfigure}
    \caption{Tracking a circular inspection trajectory around one tree while the quadrotor $x$ axis points toward the center of the tree.}
    \label{fig:inspection_circular}
\end{figure*}

Once the desired location is reached, the quadrotor must track a circumference around one trunk, maintaining a horizontal distance of 50 cm from the trunk and at a constant altitude, pointing its view-pose (positive body $x$ axis) toward the trunk center. Figure \ref{fig:inspection_circular} shows the performance of the RL-based controller in a circular inspection behavior.
\subsubsection{Track the helix to scan the trunk}
\begin{figure*}[h!]
    \centering
    \begin{subfigure}[t]{0.33\linewidth}
        \centering
        \includegraphics[width=0.8\textwidth]{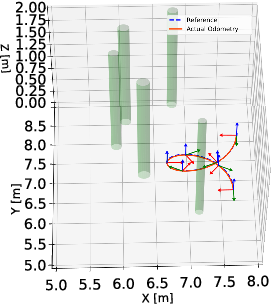}
    \end{subfigure}
    \hfill
    \begin{subfigure}[t]{0.62\linewidth}
        \centering
        \includegraphics[width=0.9\textwidth]{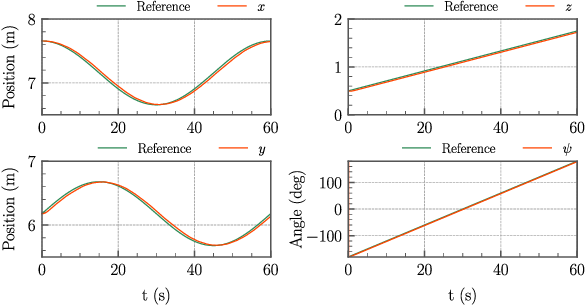}
    \end{subfigure}
    \caption{The quadrotor inspects one tree in a helix trajectory by varying its yaw orientation to see toward the tree in the whole trajectory.
    }
    \label{fig:inspection_helix}
\end{figure*}

Around a chosen tree, the quadrotor must track a helix trajectory around the trunk at a safety horizontal distance of 50 cm, whereas its positive $x$ axis points its view-pose toward the trunk center during the trajectory. Figure \ref{fig:inspection_helix} presents the inspection behavior achieved by the RL-based low-level controller.

In all test scenarios, the controller follows the planned inspection trajectories with smooth and accurate motion. The small position and yaw tracking errors support the suitability of the proposed RL-based controller as the execution layer of planner-guided navigation and inspection architectures. The forest navigation scenario highlights the trained RL-based low-level quadrotor controller’s reliability and suitability for deployment as the lowest-level controller in a forest navigation stack.

\section{CONCLUSIONS} \label{sec:conclusions}
This paper presents an RL-based low-level controller for quadrotors and evaluated its integration within a planner-guided inspection architecture across five forest-like under-canopy navigation scenarios. The results showed that the proposed controller can accurately execute collision-free trajectories and inspection poses generated by the high-level planning layers, while directly commanding the motor RPMs with smooth and precise control actions. These results support the feasibility of using the proposed RL-based controller as the execution layer of autonomous navigation and inspection stacks. Future work will focus on sim-to-real transfer and real-hardware validation to assess the controller's applicability in practical aerial inspection missions in under-canopy forest environments.


%%%%%%%%%%%%%%%%%%%%%%%%%%%%%%%%%%%%%%%%%%%%%%%%%%%%%%%%%%%%%%%%%%%%%%%%%%%%%%%%

%%%%%%%%%%%%%%%%%%%%%%%%%%%%%%%%%%%%%%%%%%%%%%%%%%%%%%%%%%%%%%%%%%%%%%%%%%%%%%%%

%%%%%%%%%%%%%%%%%%%%%%%%%%%%%%%%%%%%%%%%%%%%%%%%%%%%%%%%%%%%%%%%%%%%%%%%%%%%%%%%
%\section*{APPENDIX}

%\section*{ACKNOWLEDGMENT}

\bibliographystyle{ieeetr}
\bibliography{references}

@article{kaufmann_champion-level_2023,
	title = {Champion-level drone racing using deep reinforcement learning},
	volume = {620},
	copyright = {2023 The Author(s)},
	issn = {1476-4687},
	url = {https://www.nature.com/articles/s41586-023-06419-4},
	doi = {10.1038/s41586-023-06419-4},
	language = {en},
	number = {7976},
	urldate = {2026-01-27},
	journal = {Nature},
	publisher = {Nature Publishing Group},
	author = {Kaufmann, Elia and Bauersfeld, Leonard and Loquercio, Antonio and Müller, Matthias and Koltun, Vladlen and Scaramuzza, Davide},
	month = aug,
	year = {2023},
	pages = {982--987},
}

@ARTICLE{Eschmann_learning_to_fly,
  author={Eschmann, Jonas and Albani, Dario and Loianno, Giuseppe},
  journal={IEEE Robotics and Automation Letters}, 
  title={Learning to Fly in Seconds}, 
  year={2024},
  volume={9},
  number={7},
  pages={6336-6343},
  doi={10.1109/LRA.2024.3396025}}

@ARTICLE{xing_multi_task,
  author={Xing, Jiaxu and Geles, Ismail and Song, Yunlong and Aljalbout, Elie and Scaramuzza, Davide},
  journal={IEEE Robotics and Automation Letters}, 
  title={Multi-Task Reinforcement Learning for Quadrotors}, 
  year={2025},
  volume={10},
  number={3},
  pages={2112-2119},
  doi={10.1109/LRA.2024.3520894}}

@ARTICLE{yu_mavrl,
  author={Yu, Hang and Wagter, ChristopheDe and de Croon, Guido C. H. E},
  journal={IEEE Robotics and Automation Letters}, 
  title={MAVRL: Learn to Fly in Cluttered Environments With Varying Speed}, 
  year={2025},
  volume={10},
  number={2},
  pages={1441-1448},
  doi={10.1109/LRA.2024.3522778}}

@Article{shi_multi_waypoint,
AUTHOR = {Shi, Delong and Shen, Jinrong and Gao, Mingsheng and Yang, Xiaodong},
TITLE = {A Multi-Waypoint Motion Planning Framework for Quadrotor Drones in Cluttered Environments},
JOURNAL = {Drones},
VOLUME = {8},
YEAR = {2024},
NUMBER = {8},
ARTICLE-NUMBER = {414},
URL = {https://www.mdpi.com/2504-446X/8/8/414},
ISSN = {2504-446X},
DOI = {10.3390/drones8080414}
}

@ARTICLE{yaw_forest,
  author={Yao, Haiyun and Liang, Xinlian},
  journal={IEEE Transactions on Geoscience and Remote Sensing}, 
  title={Autonomous Exploration Under Canopy for Forest Investigation Using LiDAR and Quadrotor}, 
  year={2024},
  volume={62},
  number={},
  pages={1-19},
  doi={10.1109/TGRS.2024.3401393}}

@INPROCEEDINGS{pybullet_drones,
  author={Panerati, Jacopo and Zheng, Hehui and Zhou, SiQi and Xu, James and Prorok, Amanda and Schoellig, Angela P.},
  booktitle={2021 IEEE/RSJ International Conference on Intelligent Robots and Systems (IROS)}, 
  title={Learning to Fly—a Gym Environment with PyBullet Physics for Reinforcement Learning of Multi-agent Quadcopter Control}, 
  year={2021},
  volume={},
  number={},
  pages={7512-7519},
  doi={10.1109/IROS51168.2021.9635857}}

@article{Lopez_PID,
title = {PID control of quadrotor UAVs: A survey},
journal = {Annual Reviews in Control},
volume = {56},
pages = {100900},
year = {2023},
issn = {1367-5788},
doi = {https://doi.org/10.1016/j.arcontrol.2023.100900},
url = {https://www.sciencedirect.com/science/article/pii/S1367578823000640},
author = {Ivan Lopez-Sanchez and Javier Moreno-Valenzuela}
}

@article{gamagedara_geometric,
    author = {Gamagedara, Kanishke and Lee, Taeyoung},
    title = {Geometric Adaptive Controls of a Quadrotor Unmanned Aerial Vehicle With Decoupled Attitude Dynamics},
    journal = {Journal of Dynamic Systems, Measurement, and Control},
    volume = {144},
    number = {3},
    pages = {031002},
    year = {2021},
    month = {11},
    issn = {0022-0434},
    doi = {10.1115/1.4052714},
    url = {https://doi.org/10.1115/1.4052714},
    eprint = {https://asmedigitalcollection.asme.org/dynamicsystems/article-pdf/144/3/031002/6803181/ds_144_03_031002.pdf},
}

@Article{gao_rrt,
AUTHOR = {Gao, Haitao and Hou, Xiaozhu and Xu, Jiangpeng and Guan, Banggui},
TITLE = {Quad-Rotor Unmanned Aerial Vehicle Path Planning Based on the Target Bias Extension and Dynamic Step Size RRT* Algorithm},
JOURNAL = {World Electric Vehicle Journal},
VOLUME = {15},
YEAR = {2024},
NUMBER = {1},
ARTICLE-NUMBER = {29},
URL = {https://www.mdpi.com/2032-6653/15/1/29},
ISSN = {2032-6653},
DOI = {10.3390/wevj15010029}
}

@article{song_optimal_vs_rl,
author = {Yunlong Song  and Angel Romero  and Matthias Müller  and Vladlen Koltun  and Davide Scaramuzza },
title = {Reaching the limit in autonomous racing: Optimal control versus reinforcement learning},
journal = {Science Robotics},
volume = {8},
number = {82},
pages = {eadg1462},
year = {2023},
doi = {10.1126/scirobotics.adg1462},
URL = {https://www.science.org/doi/abs/10.1126/scirobotics.adg1462},
eprint = {https://www.science.org/doi/pdf/10.1126/scirobotics.adg1462}}

@article{hwangbo_control_2017,
	title = {Control of a {Quadrotor} with {Reinforcement} {Learning}},
	volume = {2},
	issn = {2377-3766, 2377-3774},
	url = {http://arxiv.org/abs/1707.05110},
	doi = {10.1109/LRA.2017.2720851},
	language = {en},
	number = {4},
	urldate = {2023-12-15},
	journal = {IEEE Robotics and Automation Letters},
	author = {Hwangbo, Jemin and Sa, Inkyu and Siegwart, Roland and Hutter, Marco},
	month = oct,
	year = {2017},
	note = {arXiv:1707.05110 [cs]},
	pages = {2096--2103},
}

@INPROCEEDINGS{yu_multi_RL_low_level_control,
  author={Yu, Beomyeol and Lee, Taeyoung},
  booktitle={2024 American Control Conference (ACC)}, 
  title={Multi-Agent Reinforcement Learning for the Low-Level Control of a Quadrotor UAV}, 
  year={2024},
  volume={},
  number={},
  pages={1537-1542},
  doi={10.23919/ACC60939.2024.10644789}}

@ARTICLE{zhao_speed_adaptation,
  author={Zhao, Guangyu and Wu, Tianyue and Chen, Yeke and Gao, Fei},
  journal={IEEE Robotics and Automation Letters}, 
  title={Learning Speed Adaptation for Flight in Clutter}, 
  year={2024},
  volume={9},
  number={8},
  pages={7222-7229},
  doi={10.1109/LRA.2024.3421789}}

@ARTICLE{Malczyk_semantically,
  author={Malczyk, Grzegorz and Kulkarni, Mihir and Alexis, Kostas},
  journal={IEEE Robotics and Automation Letters}, 
  title={Semantically-Driven Deep Reinforcement Learning for Inspection Path Planning}, 
  year={2025},
  volume={10},
  number={7},
  pages={7206-7213},
  doi={10.1109/LRA.2025.3575331}}

@INPROCEEDINGS{zhang_rpms,
  author={Zhang, Dingqi and Loquercio, Antonio and Wu, Xiangyu and Kumar, Ashish and Malik, Jitendra and Mueller, Mark W.},
  booktitle={2023 IEEE International Conference on Robotics and Automation (ICRA)}, 
  title={Learning a Single Near-hover Position Controller for Vastly Different Quadcopters}, 
  year={2023},
  volume={},
  number={},
  pages={1263-1269},
  doi={10.1109/ICRA48891.2023.10160836}}

@book{aerial_workers,
	author = {Nikolakopoulos, George and Mansouri, Sina and Kanellakis, Christoforos},
	title = {Aerial {Robotic} {Workers}},
	url = {https://www.sciencedirect.com/book/edited-volume/9780128149096/aerial-robotic-workers},
	language = {en-us},
	urldate = {2026-03-19},
	journal = {ScienceDirect},
	note = {ISBN: 9780128149096},
    publisher = {Butterworth-Heinemann},
    year = {2023}}

\end{document}